# Deciding Morality of Graphs is $\mathcal{NP}$-complete

T.S. Verma
Cognative Systems Laboratory
Computer Science Department
UCLA, Los Angles, CA 90024
< verma@cs.ucla.edu >

J. Pearl
Cognative Systems Laboratory
Computer Science Department
UCLA, Los Angles, CA 90024
< judea@cs.ucla.edu >

## Abstract

In order to find a causal explanation for data presented in the form of covariance and concentration matrices it is necessary to decide if the graph formed by such associations is a projection of a directed acyclic graph (dag). We show that the general problem of deciding whether such a dag exists is $\mathcal{NP}$-complete.

## 1 INTRODUCTION

The problem addressed in this paper arises in statistical data analysis. The associations measured by statistical techniques, be they dependencies or correlations, are inherently symmetric. In contrast, the models with which analysts choose to *explain* data consist of asymmetric associations, often invoking notions of influence and causation, and normally organized in the form of a dag [5, 6, 15]. This paper assesses the complexity of deciding whether a dag explanation exists for an observed set of *strong* associations, that is, pairwise dependencies that hold when we condition on all other variables in the system.

There are several technical reasons why analysts prefer dag structures for explanatory purposes [14]. First, each parameter in the dag has a well understood meaning since it is a conditional probability, i.e., it measures the probability of a response variable given a particular configuration of explanatory (parent) variables, with all other variables unspecified. Second, the task of estimating the parameters in the dag model is extremely simple, as it can be decomposed into a sequence of local estimation analyses, each involving a variable and its parent set in the dag. Third, general results are available for reading all implied independencies directly off the dag [4, 6, 11] and for deciding from the topology of two given dags whether they specify the same set of independence-restrictions on the joint distribution [12], and whether one dag specifies all the restrictions specified by the other [8].

However, the primary reason for the ubiquity of dag models lies, we believe, with their connection to causality. Each dag describes a stepwise stochastic process by which the data *could have been* generated and in this way it may "prove the basis for developing causal explanations" [1]. Causal models, no matter how they are represented, discovered or tested, are more useful than associational models, because causal model provide information about the dynamics of the system under study, In other words, a joint distribution tells us how probable events are and how probabilities would change with subsequent observations, but a causal model also tells us how these probabilities would change as a result of external interventions in the system [7]. Such information is indispensable in most decision making applications, including policy analysis and treatment management.

It is well known that in order for a dag $D$ to represent a stable causal model compatible with an observed distribution P, all the conditional independencies embodied in $D$ must be valid in $P$ [9]. The problem of deciding whether a given list $M$ of conditional independencies can be faithfully represented by a dag was treated in [13] and was shown to require a polynomial (in $|M|$) number of steps. However, $M$ may be very large as the total number of conditional independencies can in general grow exponentially with the number of variables. Thus, it is desirable to devise a test based on more limited information which is readily available to the analyst. Following Pearl and Wermuth [10], we assume that for each pair of variables $i$ and $j$, we can measure whether $i$ and $j$ are independent given all other $n-2$ variables. In the case of normal variables, such independencies can be readily obtained from the covariance matrix, as they correspond to the zero entries in the concentration matrix (the inverse covariance matrix). When the non-zero entries are represented as edges in an undirected graph $G$, the existence of a dag model of the data entails the existence of a dag $D$ such that $G$ is the "moral" graph of $D$, namely, every pair of nodes sharing a child in $D$ are adjacent in $G$ (hence the metaphor "moral" [5]).[1]

---

[1]The reason for "marrying" the parent nodes is that when conditioning on all other variables the parents may become dependent and are thus indistinguishable from any other adjacent pair of variables—both give rise to a non-

This paper shows that the problem of deciding the existence of such a dag is $\mathcal{NP}$-complete. In other words, there is no way (unless $\mathcal{P} = \mathcal{NP}$) to improve the conditions developed in [10] so as to obtain a polynomial procedure for deciding if a given set of strong dependencies has a causal explanation in a dag. This is, of course, a worst case result; some of the procedures discussed in [10] turn out to be effective in practice.

## 2  PRELIMINARIES

**2.1 Definition.** Given a dag $D$, the *moral graph* of $D$ is the undirected graph obtained by the following two operations:

1. Connect any two parents that have a common child.
2. Remove all orientations (arrowheads).    □

**2.2 Problem.** *Given an undirected graph $G$, decide if it is a moral graph of some dag. If such a dag exists we say that $G$ is moral.*    □

**2.3 Lemma.** *$G$ is moral iff there exists an acyclic orientation of a subset $E'$ of its edges, such that every unoriented edge connects the tail of two oriented edges with a common head (two parents of some common child) and no two parents are non-adjacent.*    □

**2.4 Example.** Consider the graphs G0 and G1:

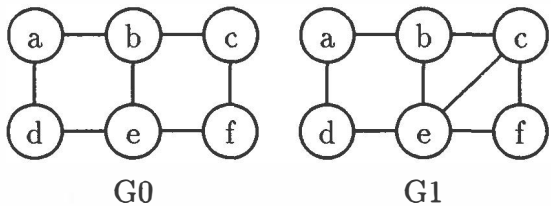

G0             G1

$G1$ is a moral graph of dag D1:

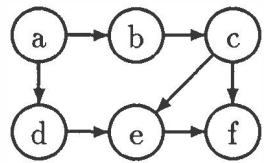

while $G0$ is not a moral graph of any dag.    □

**2.5 Definition.** A clique is said to be *exterior* if it contains at least one vertex which is adjacent only to members of that clique. Any such vertex will be called "extreme".    □

---
zero entry in the concentration matrix.

## 3  SUFFICIENT CONDITIONS FOR MORALITY

Pearl and Wermuth [10] have developed several useful sufficient conditions ranging from trivial ones to not so obvious ones.

**3.1 Proposition.** *$G$ is moral if it is chordal.*    □

**3.2 Proposition.** *$G$ is moral if every chordless n-cycle in $G$, $n \geq 4$, has at least one edge that is in some exterior clique.*    □

**3.3 Proposition.** *$G$ is moral if its maximal cliques form a web.*    □

A "web" is a collection of subsets (called components), at least one of which is exterior, such that when an exterior component is removed, the resulting structure is again, either a web or empty [2]. The examples below show that none of these sufficient conditions is necessary.

**3.4 Example.** Consider the dag $D2$ whose moral graph is $G2$:

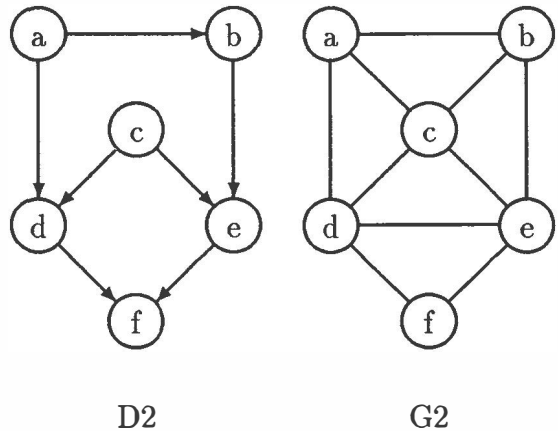

D2             G2

$G2$ is a moral graph that is not a web. The cliques are: $abc$, $acd$, $bce$, $cde$ and $def$. Only $def$ is exterior, removing $def$ leaves us with a structure that has no exterior component, hence $G2$ is not a web. Yet $G2$ is clearly the moral graph of $D2$.    □

**3.5 Example.** $G1$ in Example 2.4 violates condition 3.3, yet it is moral.    □

## 4  NECESSARY CONDITIONS FOR MORALITY

Similarly, there are several computationally attractive necessary conditions [10].

**4.1 Proposition.** *G is moral only if every chordless n-cycle, $n \geq 4$, has at least one edge that resides in some k-clique, $k \geq 3$.*  □

**4.2 Proposition.** *G is moral only if it has an exterior clique.*  □

**4.3 Example.** To see that neither condition 4.1 nor 4.2 is sufficient, consider the graph $G3$:

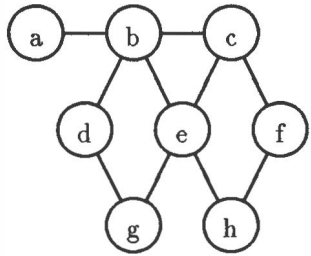

it is not moral but it satisfies both 4.1 and 4.2.  □

## 5 ANOTHER SUFFICIENT CONDITION

The following theorem provides a procedure that properly discriminates all the examples shown so far, yet it is not powerful enough to identify some moral graphs.

**5.1 Theorem.** *[10] G can be generated by a dag if all edges of G can be eliminated by repeated application of the following steps:*

*1. An exterior clique C is selected, and an extreme vertex v is identified within that clique.*

*2. A marked subgraph $G'$ is induced by removing all edges that touch v, and marking all edges in C that do not touch v.*

*3. Steps 1-2 are repeated on the induced subgraph.*

*4. If no exterior clique can be found in Step 1, then remove any marked edge and repeat Steps 1-3.*  □

We see that Proposition 3.3 is a special case of Theorem 5.1; if $G$ is a web, none of the marked edges need be removed.

The following example shows that the elimination strategy of Theorem 5.1 is not complete, that is, failure to eliminate all vertices in one ordering does not imply that no elimination ordering exists.

**5.2 Example.** Consider the graph $G4$:

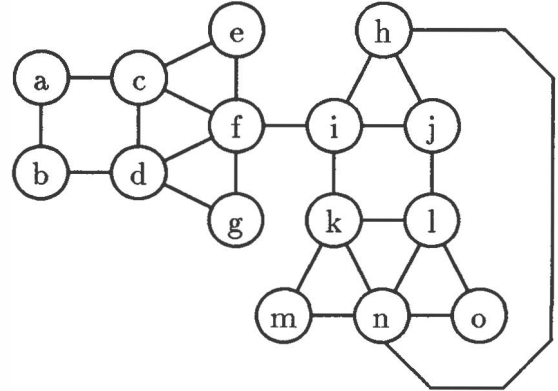

The only extreme vertices are $e$, $g$, $m$ and $o$. The result of removing these in any order is a graph with no exterior cliques and with the following marked links: $c - f, d - f, k - n$, and $l - n$. If we first remove the marked edges $f - c$ and $f - d$, the process will halt (because the cycle $a-b-c-d-a$ cannot be eliminated.) However, if we first remove the marked edges $l-n$ and $k - n$, then we will find a good elimination ordering: $..., n, h, j, l, k, i, f, ....$  □

The proof of $\mathcal{NP}$-completeness exploits the fact that it is impossible by local means to decide which of the marked edges should be removed first. While in the example above it is clear that one should postpone the removal of $f - c$ and $f - d$, because it leads to an impasse (the 4-cycle $a - b - c - d - a$), such local clues are not available in the general case.

## 6 COMPLEXITY ANALYSIS

**6.1 Theorem.** *Graph morality is $\mathcal{NP}$-complete.*

**Proof:** We first note that the problem is in $\mathcal{NP}$, because checking whether a graph $G = (V, E)$ is the moral graph of a dag $D = (V, E')$ over the same set of vertices takes $O(|V| + |E| + |E'|)$ time.

To show that the problem is $\mathcal{NP}$-complete, it is enough to show that 3-SAT is polynomially transformable into graph morality. Given an expression $F$ in 3-CNF with $n$ variables and $t$ factors it is possible to construct, in time polynomial in $n + t$, a graph $G = (V, E)$ with $32n + 22t + 8$ vertices, such that $G$ is moral if and only if $F$ is satisfiable.

The remainder of the proof consists of four parts. The first part describes the construction of the undirected graph, $G$, corresponding to the given 3-CNF expression, $F$. In the second part, it is assumed that $G$ is the moral graph of some dag $D$ and some constraints upon any such dag are derived. In the third part, it is shown that if $G$ is a moral graph then there exists a satisfying truth assignment of $F$. Finally, in the fourth part, it is shown that if $F$ is satisfiable then $G$ is a moral graph.

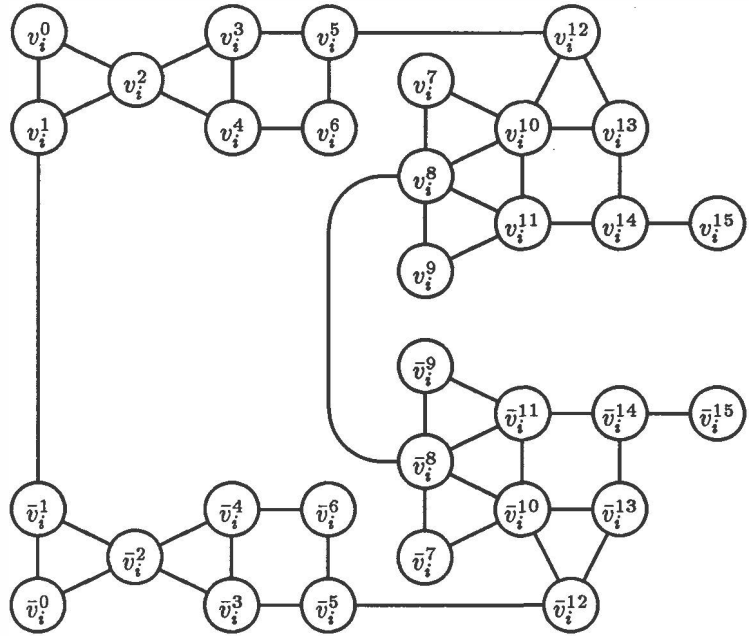

Figure 1: The subgraph corresponding to variable $v_i$.

### 6.1.1 Construction

Let $v_1, v_2, ..., v_n$ and $F_1, F_2, ..., F_t$ be the variables and factors of $F$, respectively. Let $v_i^j$, $\bar{v}_i^j$ for $1 \leq i \leq n$ and $0 \leq j \leq 15$, $F_i^j$ for $1 \leq i \leq t$ and $0 \leq j \leq 21$, and $S^j$ for $0 \leq j \leq 7$ all be new distinct symbols. Without loss of generality, assume that every variable appears as a positive literal and as a negative literal somewhere in $F$. That is, for each $1 \leq i \leq n$ there exists $1 \leq j, j' \leq t$ s.t. $v_i$ is a term of factor $F_j$ and $\bar{v}_i$ is a term of factor $F_{j'}$ since any expression in 3-CNF without this property is satisfiable and this property can be tested in linear time without use of the transformation to morality. The vertices of $G$ are:

1. $v_i^j$ and $\bar{v}_i^j$ for $1 \leq i \leq n$ and $0 \leq j \leq 15$.
2. $F_i^j$ for $1 \leq i \leq t$ and $0 \leq j \leq 21$.
3. $S^j$ for $0 \leq j \leq 7$.

The edges of $G$ are given by the following:

1. For each $1 \leq i \leq n$, the nodes $v_i^j$ and $\bar{v}_i^j$ for $0 \leq j \leq 15$ form a subgraph corresponding to variable $v_i$ as shown in Figure 1.
2. For each $1 \leq i \leq t$, the nodes $F_i^j$ for $0 \leq j \leq 21$ form a subgraph corresponding to factor $F_i$ as shown in Figure 2.
3. The nodes $S^j$ for $0 \leq j \leq 7$ form an auxiliary subgraph as shown in Figure 3.

4. The edges $(\bar{v}_i^0, v_{i+1}^0)$ for $1 \leq i < n$, connect the subgraphs corresponding to the variables forming a chain.
5. The edges $(S^0, v_1^0)$ and $(\bar{v}_n^0, S^5)$ connect the chain of subgraphs corresponding to the variables with the auxiliary subgraph.
6. The edges $(S^7, F_i^{21})$ for $1 \leq i \leq t$ connect the subgraphs corresponding to the factors with the auxiliary subgraph.

The following three classes of links inter-connect the subgraphs corresponding to the variables with the subgraphs corresponding to the factors forming a clique for every literal. The size of the clique for any literal is one more than the number of times that literal appears in $F$.

7. $(v_i^{15}, F_j^k)$ for any $1 \leq i \leq n$, $1 \leq j \leq t$ and $0 \leq k \leq 2$ such that term $(k+1)$ of factor $F_j$ is the positive literal $v_i$.
8. $(\bar{v}_i^{15}, F_j^k)$ for any $1 \leq i \leq n$, $1 \leq j \leq t$ and $0 \leq k \leq 2$ such that term $(k+1)$ of factor $F_j$ is the negative literal $\bar{v}_i$.
9. $(F_i^j, F_k^l)$ for any $1 \leq i, k \leq t$ and $0 \leq j, k \leq 2$ such that term $(j+1)$ of factor $F_i$ is the same as term $(l+1)$ of factor $F_k$.

### 6.1.2 Constraints

Suppose that there exists a dag $D = (V, E')$ such that the moral graph of $D$ is $G$.

Since $S^i$ for $1 \leq i \leq 4$ form a chordless 4-cycle, it must be the case that $D$ does not contain at least one of these edges. However, the nodes incident upon the missing edge must have a common neighbor due to Lemma 2.3. Furthermore those nodes must be ancestors of their common neighbor in $D$. Thus, $S^1$ and $S^2$ are not adjacent in $D$, while both $S^1 \to S^0$ and $S^2 \to S^0$ in $D$.

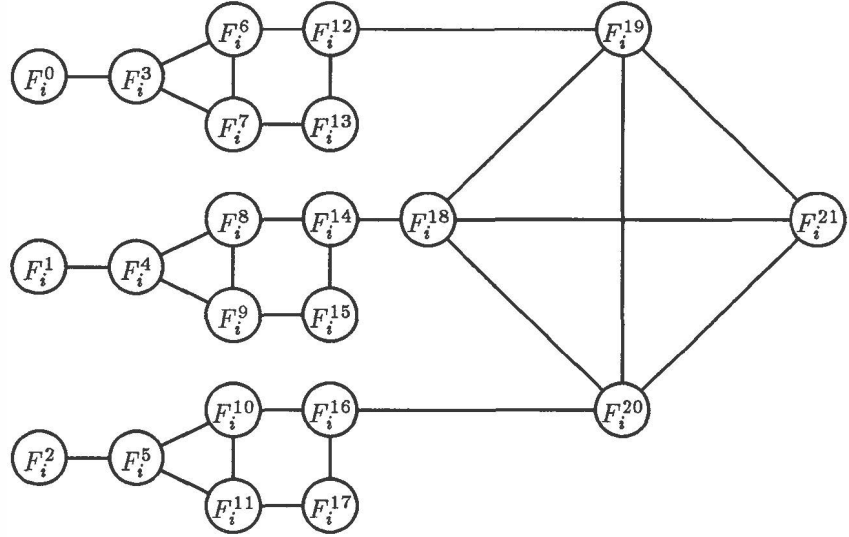

Figure 2: The subgraph corresponding to factor $F_i$.

Similarly, since $v_i^j$ for $1 \leq i \leq n$ and $3 \leq j \leq 6$ form a chordless 4-cycle, $v_i^3$ and $v_i^4$ must not be adjacent in $D$, while both $v_i^3 \to v_i^2$ and $v_i^4 \to v_i^2$ in $D$. By symmetry the same result holds for the corresponding nodes $\bar{v}_i^j$.

The edge joining $S^0$ and $v_1^0$ must appear in $D$. Now since $S^1 \to S^0$ and $S^1$ is not adjacent to $v_1^0$ it must be the case that $S^0 \to v_1^0$ in $D$. This along with the facts that $v_1^3 \to v_1^2$ and $S^0$ is not adjacent to $v_1^2$ and $v_1^3$ is not adjacent to $v_1^0$ imply that there can be no edge between $v_1^0$ and $v_1^2$ in $D$, which in turn implies that both $v_1^0 \to v_1^1$ and $v_1^2 \to v_1^1$ in $D$. Now since $v_1^0$ is not adjacent to $\bar{v}_1^1$ it follows that $v_1^1 \to \bar{v}_1^1$.

Again since $v_1^1$ is not adjacent to $\bar{v}_1^2$ and $\bar{v}_1^4$ is not adjacent to $\bar{v}_1^1$ it follows that there can be no edge between $\bar{v}_1^1$ and $\bar{v}_1^2$ in $D$, which in turn implies that both $v_1^1 \to \bar{v}_1^0$ and $\bar{v}_1^2 \to \bar{v}_1^0$ in $D$. Now since $\bar{v}_1^1$ is not adjacent to $v_2^0$ it follows that $\bar{v}_1^0 \to v_2^0$.

These last two paragraphs serve as the base step of an inductive argument that prove the following about $D$ for all $1 \leq i \leq n$:

1. If $i > 1$ then $\bar{v}_{i-1}^0 \to v_i^0$.
2. $v_i^0$ is not adjacent to $v_i^2$.
3. $v_i^0 \to v_i^1$.
4. $v_i^2 \to v_i^1$.
5. $v_i^1 \to \bar{v}_i^1$.
6. $\bar{v}_i^1$ is not adjacent to $\bar{v}_i^2$.
7. $\bar{v}_i^1 \to \bar{v}_i^0$.
8. $\bar{v}_i^2 \to \bar{v}_i^0$.
9. If $i < n$ $\bar{v}_i^0 \to v_{i+1}^0$.

The inductive part of the proof is virtually identical except for the replacement of $S^0$ by $\bar{v}_{i-1}^0$, $v_1$ by $v_i$, and $v_2$ by $v_{i+1}$.

Note that $\bar{v}_n^0 \to S^5$ also follows from the induction. Since $\bar{v}_n^0$ is not adjacent to $S^6$ it follows that $S^5 \to S^6$. Similarly, since $S^5$ is not adjacent to $S^7$ it follows that $S^6 \to S^7$. And, finally, since $S^6$ is not adjacent to $F_i^{21}$ for $1 \leq i \leq t$ it follows that $S^7 \to F_i^{21}$.

Therefore, $v_i^3$, $v_i^4$, $\bar{v}_i^3$ and $\bar{v}_i^4$ are ancestors of $S^7$ in $D$ for all $1 \leq i \leq n$. Now consider $v_i^5$ for any $1 \leq i \leq n$. Either $v_i^5 \to v_i^3$ or $v_i^3 \to v_i^5$. In the latter case $v_i^5 \to v_i^6 \to v_i^4$. Thus, either way $v_i^5$ is an ancestor of $S^7$. Similarly, $\bar{v}_i^5$ is also an ancestor of $S^7$.

Now consider $F_i^j$ for all $1 \leq i \leq t$ and $6 \leq j \leq 11$. By Lemma 2.3 the edges $F_i^6 - F_i^7$, $F_i^8 - F_i^9$ and $F_i^{10} - F_i^{11}$ cannot appear in $D$, and these pairs of nodes must be parents of $F_i^3$, $F_i^4$ and $F_i^5$, respectively. Furthermore, $D$ must also contain $F_i^3 \to F_i^0$, $F_i^4 \to F_i^1$ and $F_i^5 \to F_i^2$.

The cliques that inter-connect the subgraphs corresponding to the variables with those corresponding to the factors (edge construction rules 7–9) all have the following property: every $F$ node in the clique has a unique parent in $D$ and none of these parents are adjacent. Furthermore, the only other node in the clique is either $v_k^{15}$ or $\bar{v}_k^{15}$ for $1 \leq k \leq n$. It follows that the nodes of the clique in $G$ must form a star in $D$ where each of the $F$ nodes is a parent of the $v$ or $\bar{v}$ node, and the $F$ nodes are not adjacent to each other. In turn, this implies that $v_i^{15} \to v_i^{14}$ for all $1 \leq i \leq n$ and for $\bar{v}$.

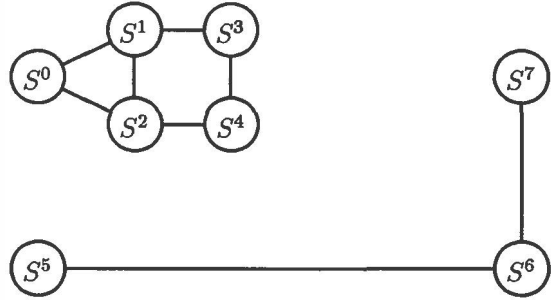

Figure 3: The auxiliary subgraph.

Now, for each $1 \leq i \leq n$ either $\bar{v}_i^8 \to v_i^8$ or $v_i^8 \to \bar{v}_i^8$ in $D$. If $\bar{v}_i^8 \to v_i^8$ then analysis of the 4-cycle $v_i^{10} - v_i^{11} - v_i^{14} - v_i^{13} - v_i^{10}$ reveals that $v_i^{15}$ would be an ancestor of $S^7$ in $D$. Similarly, if $v_i^8 \to \bar{v}_i^8$ then analysis of the corresponding $\bar{v}$ 4-cycle reveals that $\bar{v}_i^{15}$ would be an ancestor of $S^7$ in $D$.

The key to the analysis (for $\bar{v}_i^8 \to v_i^8$) is that $v_i^{10} - v_i^{11}$ and $v_i^{10} - v_i^{13}$ can not both appear in $D$. In the later case, it is trivial to demonstrate that $D$ would contain $v_i^{13} \to v_i^{12} \to v_i^5$ and hence $v_i^{15}$ would be is an ancestor of $S^7$. To conclude the analysis, note that the former case is impossible since if $v_i^{10} - v_i^{13}$ does not appear in $D$ it follows that both nodes must be parents of $v_i^8$, but $\bar{v}_i^8$ is also a parent of $v_i^8$ and thus $G$ would not be the moral graph of $D$ as $\bar{v}_i^8$ is not adjacent to either of the nodes, $v_i^{10}$ or $v_i^{13}$, in $G$.

Next, an analysis similar to the one for the cycle involving $v_j^3$, $v_j^4$, $v_j^5$ and $v_j^6$ will reveal that, for all $1 \leq i \leq t$, $F_i^{12}$ must be an ancestor of $F_i^0$ in $D$. Correspondingly, $F_i^{14}$ is an ancestor of $F_i^1$ and $F_i^{16}$ is an ancestor of $F_i^2$.

Finally, an analysis of the cliques $F_i^{18} F_i^{19} F_i^{20} F_i^{21}$, for all $1 \leq i \leq t$, reveals that, in $D$, $F_i^{21}$ must be the parent of one of the other three nodes (of this clique in $G$), because not all three links can be removed in $D$, and $F_i^{21}$ is the child of $S_7$ and $S_7$ is not adjacent to the other three nodes (of this clique in $G$). Therefore, it is easy to show that $F_i^{21}$ is an ancestor of $F_i^0$ or $F_i^1$ or $F_i^2$. Note that each of these are ancestors of $v_j^{15}$ or $\bar{v}_j^{15}$ for some $1 \leq j \leq n$.

At this point, $D$ is almost constrained to contain a directed cycle. For each $1 \leq i \leq t$ and each $1 \leq j \leq n$, either $v_j^{15}$ or $\bar{v}_j^{15}$ is an ancestor of $F_i^{21}$, and $F_i^{21}$ must be an ancestor of at least one of three (the particular three are defined by the graph) $v_j^{15}$ or $\bar{v}_j^{15}$ nodes.

### 6.1.3   Morality implies Satisfiability

If there is a dag $D$ such that $G$ is the moral graph of $D$, then the following truth assignment satisfies $F$: let $v_i = True$ iff $v_i^8 \to \bar{v}_i^8$.

### 6.1.4   Satisfiability implies Morality

To get from a satisfying truth assignment to a dag, first remove and direct the necessary arcs as described above. Then direct $v_j^8 \to \bar{v}_j^8$ iff $v_j = True$. Next, make $F_i^{21}$ a parent of any (or all) of the nodes $F_i^{18}$, $F_i^{19}$, $F_i^{20}$ which are ancestors of $v_j^{15}$ or $\bar{v}_j^{15}$ nodes that are not ancestors of $F_i^{2}1$, i.e., that correspond to true literals. The other arc(s) in this clique should be removed in $D$.

The remaining arcs can be easily directed without conflicts. The only interesting part is directing the links connecting the nodes $v_i^j$ for $1 \leq i \leq n$ and for $7 \leq j \leq 15$ and for $\bar{v}$. The direction of these arcs depends upon the direction of the arc between $v_i^8$ and $\bar{v}_i^8$ and can be seen in the following example. $\square$

**6.2 Example.** Consider the 3-SAT problem $F = (X + Y + Z)(\bar{X} + \bar{Y} + Z)(\bar{X} + \bar{Y} + \bar{Z})$. The undirected skeleton (including dashed links) of the graph in Figure 4 represents $G$, the graph corresponding to $F$. The directions and dashed links represent the constraints placed upon any dag $D$ s.t. $G$ is the moral graph of $D$. The dag in Figure 5 corresponds to the truth assignment: $X = true$, $Y = false$, $Z = false$.

$\square$

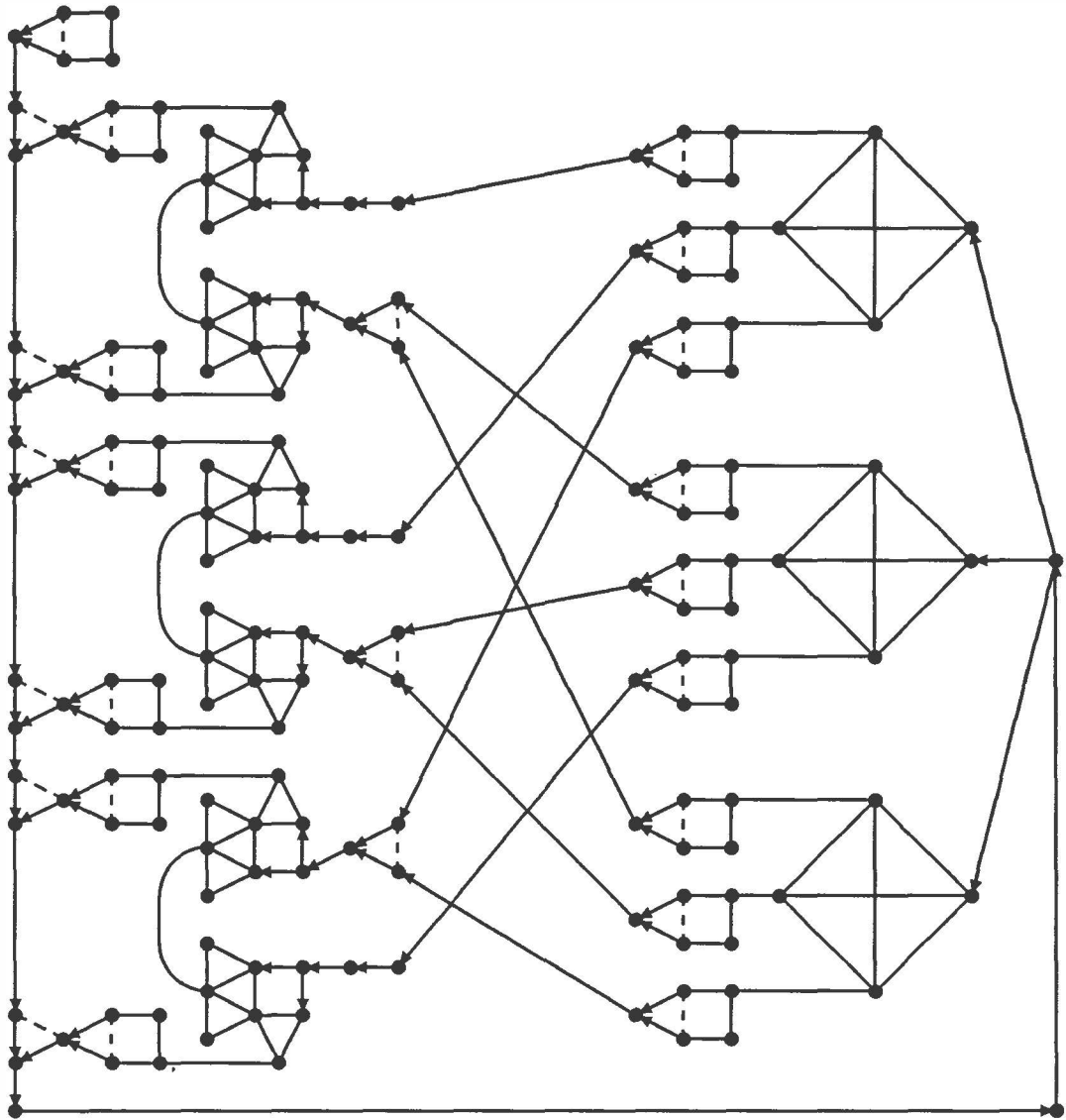

Figure 4: Graph and Constraints corresponding to $F = (X + Y + Z)(\bar{X} + \bar{Y} + Z)(\bar{X} + \bar{Y} + \bar{Z})$.

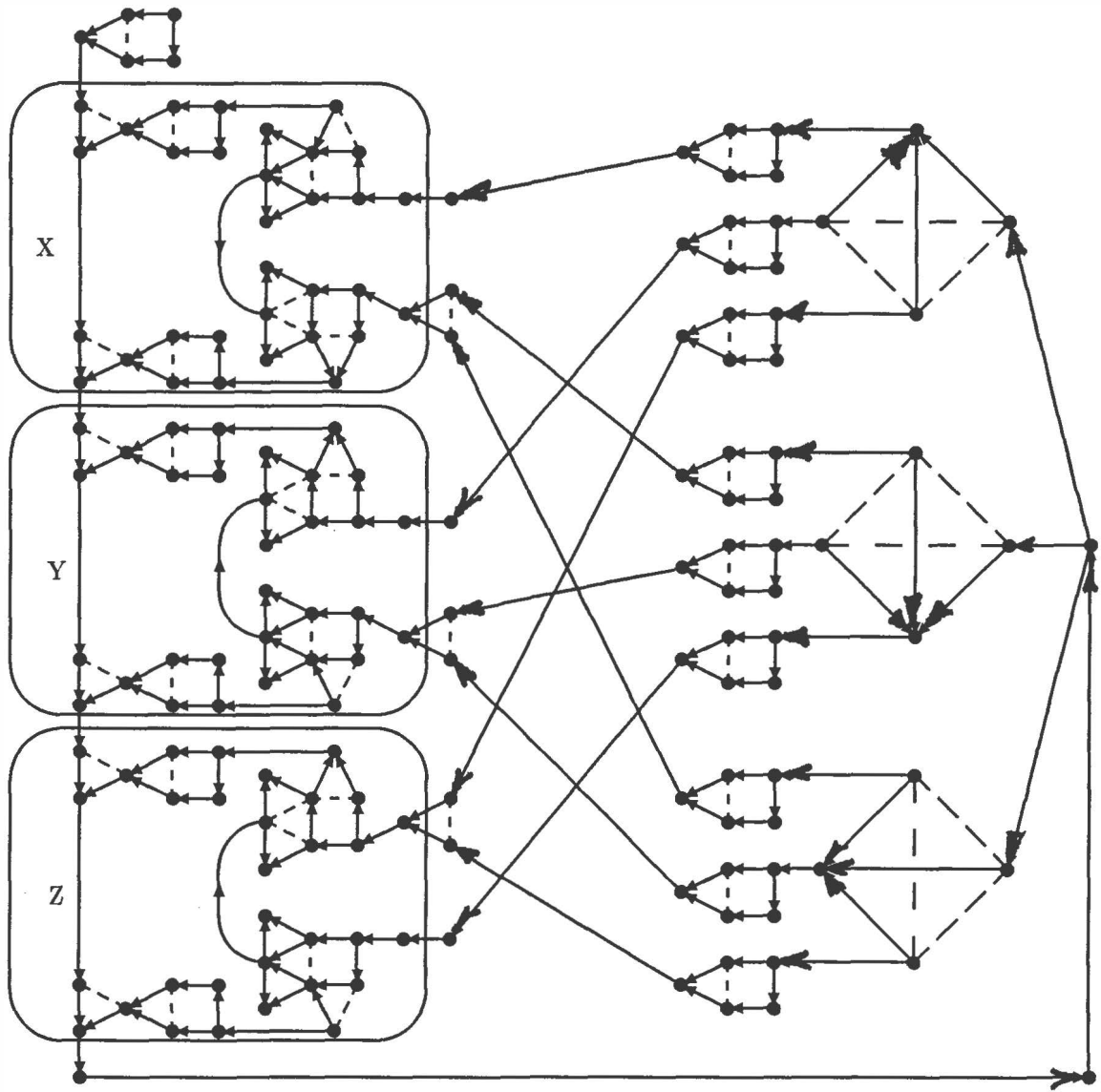

Figure 5: A DAG corresponding to $X = true$, $Y = false$, $Z = false$.